\documentclass[10pt,twocolumn,letterpaper]{article}

\usepackage[pagenumbers]{cvpr} 

\definecolor{cvprblue}{rgb}{0.21,0.49,0.74}
\usepackage[pagebackref,breaklinks,colorlinks,allcolors=cvprblue]{hyperref}

\def\paperID{*****} 
\def\confName{CVPR}
\def\confYear{2025}

\title{EgoVIS@CVPR: What Changed and What Could Have Changed? State-Change Counterfactuals for Procedure-Aware Video Representation Learning}

\author{
Chi-Hsi Kung$^*$ \\
Indiana University 
\and
Frangil Ramirez$^*$ \\
Indiana University 
\and
Juhyung Ha \\
Indiana University 
\and
Yi-Ting Chen$^\dagger$ \\
National Yang-Ming Chiao-Tung University 
\and
David Crandall$^\dagger$ \\
Indiana University 
\and
Yi-Hsuan Tsai$^\dagger$ \\
Atmanity Inc. 
}

\begin{document}

        

\twocolumn[{%
    \renewcommand\twocolumn[1][]{#1}%
    \maketitle
    \begin{center}
        \centering
        \captionsetup{type=figure}
        \includegraphics[width=0.75\textwidth, height=0.3\textwidth]{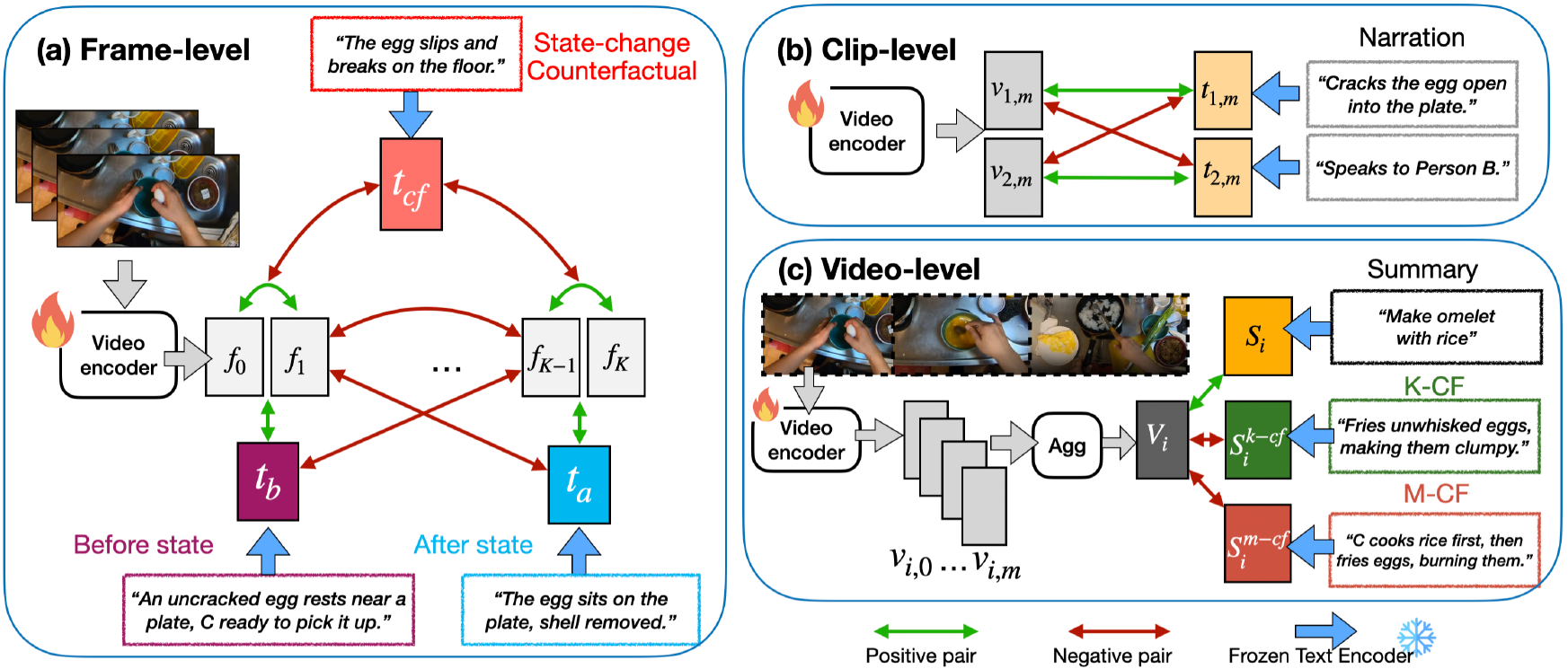} 
        \captionof{figure}{
            Illustration of our learning framework. (a) We train \textbf{frame} features by temporally contrasting neighboring and distant frames by incorporating \textcolor{RedViolet}{Before}, \textcolor{cyan}{After}, and \textcolor{red}{State-change Counterfactuals (SC-CF)}. (b) We align \textbf{clip} features with narration features. (c) Clip features are aggregated with the aggregator (Agg) as \textbf{video} features. Then the video features are contrasted positively with summaries and negatively with \textcolor{OliveGreen}{Missing-step Counterfactuals (K-CF)} and \textcolor{BrickRed}{Misordered Counterfactuals (M-CF)}. Note that all text features are extracted with the frozen text encoder.
            }
    \label{fig:teaser}
\end{center}%
}]

\def\thefootnote{*}\footnotetext{Equal contribution. $\dagger$Equal advising.}
\begin{abstract}
Understanding a procedural activity requires modeling both how action steps transform the scene, and how evolving scene transformations can influence the sequence of action steps, even those that are accidental or erroneous. Yet, existing work on procedure-aware video representations fails to explicitly learned the state changes (scene transformations). In this work, we study procedure-aware video representation learning by incorporating state-change descriptions generated by LLMs as supervision signals for video encoders. Moreover, we generate state-change counterfactuals that simulate hypothesized failure outcomes, allowing models to learn by imagining the unseen ``What if'' scenarios. This counterfactual reasoning facilitates the model's ability to understand the cause and effect of each step in an activity.
To verify the procedure awareness of our model, we conduct extensive experiments on procedure-aware tasks, including temporal action segmentation, error detection, and more. Our results demonstrate the effectiveness of the proposed state-change descriptions and their counterfactuals, and achieve significant improvements on multiple tasks. 
Full paper~\cite{kung2025changed} is available at
\href{https://arxiv.org/abs/2503.21055}{https://arxiv.org/abs/2503.21055}.
\end{abstract}    
\section{Introduction}
\label{sec:intro}

%
Understanding procedure activities from video data is essential for a variety of applications, including video retrieval~\cite{Wray_2021_CVPR}, 
intelligent collaborative agents~\cite{jenamani2024feel}, and robot learning from human demonstration~\cite{ren2025motion}. 
In contrast to general video action recognition that focuses on a single step in a short clip~\cite{tong2022videomae,Kung_2024_CVPR}, procedure-aware video understanding requires capturing both the ``what changed'' (actual action-induced state transformations)~\cite{souvcek2023genhowto}, such as \textcolor{RedViolet}{Before} and \textcolor{cyan}{After} state shown in Figure~\ref{fig:teaser} (a), and ``what could have changed'' (hypothetical deviations)~\cite{zhang2021if}, such as \textcolor{red}{State-change counterfactuals} shown in Figure~\ref{fig:teaser} (a) and \textcolor{OliveGreen}{Missing-step} and \textcolor{BrickRed}{Misordered} shown in Figure~\ref{fig:teaser} (c).
These capabilities are key to understanding long-form procedures with sequentially dependent steps.

%
%

        

%
There have been many approaches proposed to learn procedure-aware representations, including learning spatiotemporal features~\cite{fan2021multiscale,tong2022videomae}, using action labels as supervision~\cite{xiao2022hierarchical,zhong2023learning}, incorporating temporal order of steps~\cite{zhong2023learning}, and consulting external activity procedure knowledge databases~\cite{zhou2023procedure}.
However, these methods often fail to model how scene states evolve or could have evolved under different action outcomes.
We propose a hierarchical video representation learning framework for procedure understanding that leverages state changes and counterfactuals generated by an LLM~\cite{chen2020simple}. At the clip level, we model \emph{before}, \emph{after}, and \emph{counterfactual} states to capture local action-induced changes. 
We use temporal contrastive learning to bring \emph{after} states close to later frames' features and push apart \emph{before} and counterfactuals. At the video level, we extend this with long-form counterfactuals like \emph{missing-step} and \emph{misordered}, improving procedure awareness.

We evaluate the learned procedure-aware video representations in three key procedural video understanding tasks---error detection, temporal action segmentation, and action phase classification and frame retrieval---and show that they significantly enhance performance compared to strong baselines. 
%
In summary, our contributions are:
\begin{enumerate}
   \item A video representation method leveraging state changes and counterfactuals for procedural understanding.

    \item A hierarchical framework aligning frame-, clip-, and video-level features using these descriptions.

    \item Our method achieves state-of-the-art results on procedure-aware tasks with detailed analysis.
\end{enumerate}

\section{Related Work}
\label{sec:related_work}

Recently, several papers have attempted to align video features in datasets such as HowTo100M~\cite{miech2019howto100m} or Ego4D~\cite{grauman2022ego4d} 
with text descriptions extracted through
ASR~\cite{miech2020end,mavroudi2023learning}, refined subtitles~\cite{lin2022learning} with the external database WikiHow~\cite{koupaee2018wikihow,mavroudi2023learning}, or manually labeled annotations~\cite{kevin2022egovlp,zhong2023learning,ashutosh2023hiervl}. 
%
However, these approaches fail to explicitly learn action-induced state changes or detect deviations like erroneous steps, leading to overfitting on correctly executed actions and limiting procedural understanding.
We address this by incorporating state-change descriptions and counterfactuals to model action transformations and hypothetical failures for improved procedure awareness.

        

Visual state changes have been widely studied~\cite{liang2022var,souvcek2022look,niu2024schema}, and we draw inspiration from SCHEMA~\cite{niu2024schema} in using LLMs to generate before- and after-state descriptions.
Our work differs in three key ways: (1) we target general video representation learning beyond a single task; (2) we incorporate counterfactuals for causal reasoning; and (3) our descriptions capture state changes of objects, humans, and environments---not just interacted objects.


\section{Pretraining Objective: \\ State Change \& Counterfactual}
\label{sec:methods}
Here, we present our pretraining strategy that incorporates the generated state changes and state-change counterfactuals to learn procedure-aware representations.
We build upon HierVL~\cite{ashutosh2023hiervl}, a framework that learns hierarchical video-language representations at two temporal scales, clip-level and video-level. Please refer to~\cite{ashutosh2023hiervl} for an overview of the framework.
We extend HierVL with finer-grained frame-level alignment.

\noindent\textbf{Frame-Level Alignment }
At the frame level, the model is supervised by our proposed \textcolor{RedViolet}{Before state} loss $\mathcal{L}_{\text{before}}$ and \textcolor{cyan}{After-state} loss $\mathcal{L}_{\text{after}}$, enhancing action-induced transformation for procedural understanding. The mathematical formulation of each is \cite{10.5555/3495724.3497291},
\begin{align}\label{eq:frame_loss}
    \mathcal{L} &= \frac{1}{|B|} \sum_{i \in B} \frac{-1}{|P(i)|} \sum_{p \in P(i)} \log \frac{\exp(f_i^Tz_p/\tau)}{\sum_{n \in N(i)}\exp(f_i^Tz_n/\tau)},
\end{align}
where $B$ is the batch size, $f_i$ is the visual embedding of the $i^{\text{th}}$ frame, $z_j$ is either a visual or text embedding, $\tau$ is a temperature hyperparameter, and $P(i)$ and $N(i)$ denote the positive and negative samples of the $i^{\text{th}}$ frame, respectively.
Given a set of $K=4$ frames sub-sampled from a video-clip, $L_{\text{before}}$ aims to align 
earlier-in-time frames along with the \emph{before-state} text embeddings, while pushing them apart from later-in-time frames, the \emph{after-state}, and counterfactual text embeddings.
%
%
On the other hand, $L_{\text{after}}$ aims to align later-in-time frames to the after state while separating them from earlier-in-time frames, the before-state, and counterfactual text embeddings. 

\noindent\textbf{Clip-Level Alignment }
At the clip-level, we leverage the $L_{v2t}$ loss described in \cite{ashutosh2023hiervl, kevin2022egovlp}, which seeks to align the \emph{video-clip} embeddings to their corresponding text narrations from EgoClip \cite{kevin2022egovlp}. Note that since we do not train a text encoder, the symmetric $L_{t2v}$ loss is neglected here. 
For more details on this loss, please see \cite{ashutosh2023hiervl, kevin2022egovlp}. 
The resulting loss for the first two scales is therefore
\begin{align}\label{eq:child_loss}
    \mathcal{L}_{\text{child}} = \mathcal{L}_{v2t} + \lambda(\mathcal{L}_{\text{before}} + \mathcal{L}_{\text{after}}),
\end{align}
where $\lambda$ is a hyperparameter controlling the strength of the state-change aware supervision.

\noindent\textbf{Video-Level Alignment } This loss   aligns video-level visual embeddings to summary text embeddings and  enhances procedural awareness using video-level counterfactuals. 
We first obtain video-level visual embeddings from clip-level embeddings by using a self-attention block as an aggregator function. 
Then, using contrastive learning, each visual embedding is aligned to its corresponding summary text embedding and contrasted against text embeddings of the generated counterfactual \textcolor{BrickRed}{Misordered} and \textcolor{OliveGreen}{Missing-step},
%
\begin{align}\label{eq:parent_loss}
    \mathcal{L}_{\text{parent}} &= -\sum_{i \in B} \log 
    \frac{\sum_{p \in P(i)} \exp(V_i^T S_p)} 
    {\sum_{n \in N(i)}\exp(V_i^T S_n) + \exp(V_i^T S_{n,w}^{cf})},
\end{align}
where $V_i$ is the aggregated video-level visual embedding, $S_j$ is a summary text embedding and $S_{j,w}^{cf}$ are \textcolor{BrickRed}{Misordered} and \textcolor{OliveGreen}{Missing-step} counterfactual text embeddings where $w \in \{ 1, \ldots, W\}$ is the total number of counterfactuals used, and $P(i)$ and $N(i)$ denote the positive and negative samples of the $i^{\text{th}}$ video, respectively. We perform video-level alignment as in Eq. \eqref{eq:parent_loss} after every 5 mini-batches of clip- and frame-level alignment as in Eq. \eqref{eq:child_loss}. 
\section{Experiments}
\label{sec:experiments}
\noindent \textbf{Tasks}
We evaluate our model's learned representations on four procedural video-understanding tasks: temporal action segmentation on GTEA~\cite{gtea} and EgoPER~\cite{egoper}, error detection on EgoPER~\cite{egoper}, and action phase classification and frame retrieval on  Align-Ego-Exo~\cite{xue2023learning}. We report F1 score, mean average precision (mAP), edit distance, frame accuracy, or Error Detection Accuracy (EDA) metrics based on each task convention.

\noindent \textbf{Implementation Details}
We use ASFormer~\cite{chinayi_ASformer}, EgoPED~\cite{egoper}, an SVM, and nearest-neighbors for each task, respectively, and equip them with different input representation features. In the Align-Ego-Exo dataset, we merge the validation and test sets for more robust results.

\noindent \textbf{Baselines}
We compare to I3D and CLIP features~\cite{carreira2017quo,radford2021learning} commonly used in long-form video tasks and procedure-aware representations with publicly available pretrained model weights, including MIL-NCE~\cite{miech2020end}, and PVRL~\cite{zhong2023learning}. 
In addition, we evaluate the VLM HierVL~\cite{ashutosh2023hiervl} with only its video encoder.

\subsection{Experimental Results}
\noindent \textbf{Temporal Action Segmentation}
In Table~\ref{table:tsa}, we find that MIL-NCE performs significantly worse than others, which we hypothesize is due to the poor quality of description made by ASR. 
Furthermore, we compare our model with the VLM HierVL.
Even though HierVL is trained with both video and text encoders, which require considerably more computational resources, our model outperforms HierVL in most metrics on both datasets. 
The performance gap against all non-VLM models is even larger.

\noindent \textbf{Error Detection}
In Table~\ref{table:error_detection}, we observe that all procedure-aware representations outperform general visual representations, such as I3D and CLIP, highlighting the importance of procedure awareness in the context of error detection.
Our proposed method surpasses the state-of-the-art in the EDA metric~\cite{egoper}, even outperforming the VLM HierVL. 

\noindent \textbf{Action Phase Classification \& Frame Retrieval}
In Table~\ref{table:ae2} and Table~\ref{table:retrieval}, our learned representations outperform other models on average across actions in both phase classification and retrieval, on both the \emph{ego+exo views} (regular) and \emph{egocentric only} (ego) settings, while also achieving superior or competitive performance on most actions individually.

\begin{table}[t]
\centering
\scriptsize
\caption{Temporal action segmentation results on the GTEA~\cite{gtea} and EgoPER~\cite{egoper} datasets.}
\vspace{-2mm}
\resizebox{1\linewidth}{!}{
    \begin{tabular}{@{}l|ccccc|ccc@{}}
        \toprule
        \multicolumn{1}{c}{}
        & \multicolumn{5}{c}{\textbf{GTEA}} & \multicolumn{3}{c}{\textbf{EgoPER}} \\
         \cmidrule(lr){2-6} \cmidrule(lr){7-9}
        \textbf{Method} & F1@10 & F1@25 & F1@50 & Edit & Acc & F1@50 & Edit & Acc \\
        \midrule
        I3D~\cite{carreira2017quo} & \underline{90.1} & \underline{88.8} & 79.2 & 84.6 & \underline{79.7} 
        & 48.8 & 71.9 & 73.9\\
        CLIP~\cite{radford2021learning} & 88.5 & 86.2 & 77.6 & \textbf{87.1} & 75.6
        & 44.2 & 71.2 & 70.8\\
        MIL-NCE~\cite{miech2020end} & 67.9 & 61.3 &	44.6 & 67.9 & 58.3
        & 47.3 & 69.1 & 73.6\\
        PVRL~\cite{zhong2023learning}& 85.2 & 82.6 & 72.2 & 81.1 & 71.2
        & 45.6 & 73.2 & 73.4\\
        \midrule
        HierVL~\cite{ashutosh2023hiervl}& \textbf{90.4}	& 88.5 & \underline{81.2} & 86.7& 78.5
        & \underline{52.6}	& \underline{73.0} & \underline{77.3}\\
        \midrule
        Ours & 89.8 & \textbf{89.1} & \textbf{81.6} & \underline{86.8} & \textbf{80.0}
        & \textbf{54.4}	& \textbf{74.1} & \textbf{79.0}\\
        \bottomrule
    \end{tabular}
}
\label{table:tsa}
\end{table}


\begin{table}[t]
\centering
\small
\caption{Error detection results on EgoPER~\cite{egoper}. HTM denotes the HowTo100M~\cite{miech2019howto100m} dataset. \textbf{\textcolor{red}{Text}} denotes the VLM model with a trainable text encoder. The metric is EDA.
}
\vspace{-2mm}
\resizebox{1\linewidth}{!}{
        \begin{tabular}
            {@{}l@{\;}c c c  c @{\;} c@{\;} c @{\;} c @{\;} c @{\;}c @{\;}c @{\;}c@{\;}|c@{\;}c }
            \toprule
            \multicolumn{1}{l}{Method}& 
            \multicolumn{1}{l}{Pretraining Data}& 
            \multicolumn{1}{c}{Quesadilla}  & 
            \multicolumn{1}{c}{Oatmeal}  & 
            \multicolumn{1}{c}{Pinwheel}  & 
            \multicolumn{1}{c}{Coffee} &
            \multicolumn{1}{c}{Tea} &
            \multicolumn{1}{c}{All} &
             \\
             \midrule
              \textcolor{gray!80}{Random} & \textcolor{gray!80}{-} & \textcolor{gray!80}{19.9} & \textcolor{gray!80}{11.8} & \textcolor{gray!80}{15.7} & \textcolor{gray!80}{8.20} & \textcolor{gray!80}{17.0} & \textcolor{gray!80}{14.5}
             \\
             \midrule
              I3D~\cite{carreira2017quo} & Kinetics & 62.7 & 51.4 & 59.6 & 55.3 & 56.0 & 57.0
             \\
              CLIP~\cite{radford2021learning} & WIT~\cite{srinivasan2021wit}+\textbf{\textcolor{red}{Text}} & 77.6 & 69.6 & \underline{66.9} & \textbf{68.5} & {75.6} & \underline{71.6}
             \\
              MIL-NCE~\cite{miech2020end} & HTM & 77.3 & 69.8 & 65.7 & 68.0 & 69.8 & 70.1
             \\
             PVRL~\cite{zhong2023learning} & HTM & 75.7 & \underline{71.2} & 65.5 & 67.5 & \underline{76.4} & 71.3
            \\
            \midrule
             HierVL~\cite{ashutosh2023hiervl} & Ego4D+\textbf{\textcolor{red}{Text}} & \underline{77.9} & 70.8 & 65.2 & 67.4 & {75.1} & 71.3
            \\
            \midrule
             Ours & Ego4D & \textbf{78.9} & \textbf{71.6} & \textbf{68.3} & \underline{68.3} & \textbf{76.6} & \textbf{72.7}
            \\
             \bottomrule
        \end{tabular}
}
\label{table:error_detection}
\end{table}


\begin{table}[t]
\scriptsize
\caption{Action phase classification results on the Align-Ego-Exo dataset~\cite{xue2023learning}. The metric is F1 score.}
\vspace{-2mm}
\resizebox{1\linewidth}{!}{
    \begin{tabular}{@{}l|cccccccc|cc@{}}
        \toprule
         \multicolumn{1}{c}{{}}
        & \multicolumn{2}{c}{{Break Eggs}} & \multicolumn{2}{c}{{Pour Milk}} & \multicolumn{2}{c}{{Pour Liquid}} & \multicolumn{2}{c}{{Tennis Forehand}} & \multicolumn{2}{c}{{All}}\\
        {Method} & regular & ego & regular & ego & regular & ego & regular & ego & regular & ego \\
        \midrule
        CLIP~\cite{radford2021learning} & 50.1 & 54.9 & \underline{50.4} & \textbf{49.8} & 61.3 & 63.7 & \textbf{76.3} & \textbf{78.2} & \underline{59.5} & 61.6\\
        MIL-NCE~\cite{miech2020end} & 45.5 & 45.0 & 45.9 & 44.2 & 61.2 & 65.3 & 59.5 & 62.3 & 53.0 & 54.2\\
        PVRL~\cite{zhong2023learning} & \underline{54.6} & \underline{60.6} & \textbf{51.6} & 46.6 & \underline{63.0} & \underline{69.0} & 68.2 & 74.5 & 59.4 & \underline{62.7}\\
        \midrule
        Ours & \textbf{56.2} & \textbf{65.8} & 48.1 & \underline{47.6} & \textbf{68.1} & \textbf{70.6} & \underline{72.7} & \underline{75.1} & \textbf{61.3} & \textbf{64.8} \\
        \bottomrule
    \end{tabular}
}
\label{table:ae2}
\end{table}


\begin{table}[t]
\scriptsize
\caption{Frame retrieval results on the Align-Ego-Exo dataset~\cite{xue2023learning}. The metric is mAP@10.}
\vspace{-2mm}
\resizebox{1\linewidth}{!}{
    \begin{tabular}{@{}l|cccccccc|cc@{}}
        \toprule
         \multicolumn{1}{c}{{}}
        & \multicolumn{2}{c}{{Break Eggs}} & \multicolumn{2}{c}{{Pour Milk}} & \multicolumn{2}{c}{{Pour Liquid}} & \multicolumn{2}{c}{{Tennis Forehand}} & \multicolumn{2}{c}{{All}}\\
        {Method} & regular & ego & regular & ego & regular & ego & regular & ego & regular & ego \\
        \midrule
        CLIP~\cite{radford2021learning} & \underline{63.5} & \underline{68.0} & \textbf{59.3} & \underline{59.2} & {55.9} & 56.1 & \underline{79.1} & {88.7} & \underline{64.4} & \underline{68.0}\\
        MIL-NCE~\cite{miech2020end} & 58.0 & 57.4 & {47.3} & 51.0 & \underline{57.7} & \underline{59.2} & 74.8 & 84.3 & 59.5 & 63.0\\
        PVRL~\cite{zhong2023learning} & {59.5} & {63.1} & \underline{58.3} & \textbf{59.3} & {50.2} & {55.1} & {78.3} & \textbf{88.9} & {61.6} & {66.6}\\
        \midrule
        Ours & \textbf{66.5} & \textbf{69.4} & 51.4 & {54.9} & \textbf{62.4} & \textbf{67.8} & \textbf{79.4} & \textbf{88.9} & \textbf{64.9} & \textbf{70.3} \\
        \bottomrule
    \end{tabular}
}
\vspace{-12pt}
\label{table:retrieval}
\end{table}
\section{Conclusions}
\label{sec:conclusion}
We present a novel procedure-aware video representation learning framework that first incorporates state-change descriptions and state-change counterfactuals in clip-level alignment, enhancing causal reasoning of action transformations. 
Then, it utilizes video-level counterfactuals that perturb the local actions and create hypothesized scenarios to facilitate the understanding of activity procedures. 
Our learned representations demonstrate strong effectiveness in terms of procedure awareness and achieve state-of-the-art results on several benchmarks. 

\textbf{Acknowledgments.}
CK, FR, JH, DC were supported in part by  NSF   (DRL-2112635 via AI Institute for Engaged Learning) and  Lilly Endowment via  PTI. YC was supported in part by  National Science and Technology Council  (113-2628-E-A49-022, 114-2628-E-A49-007),   NYCU Higher Education Sprout Project, and   Ministry of Education Yushan Fellow Program Administrative Support Grant. 
{
    \small
    \bibliographystyle{ieeenat_fullname}
    \bibliography{main}

\begin{thebibliography}{32}
\providecommand{\natexlab}[1]{#1}
\providecommand{\url}[1]{\texttt{#1}}
\expandafter\ifx\csname urlstyle\endcsname\relax
  \providecommand{\doi}[1]{doi: #1}\else
  \providecommand{\doi}{doi: \begingroup \urlstyle{rm}\Url}\fi

\bibitem[Ashutosh et~al.(2023)Ashutosh, Girdhar, Torresani, and Grauman]{ashutosh2023hiervl}
Kumar Ashutosh, Rohit Girdhar, Lorenzo Torresani, and Kristen Grauman.
\newblock Hiervl.
\newblock In \emph{CVPR}, 2023.

\bibitem[Carreira and Zisserman(2017)]{carreira2017quo}
Joao Carreira and Andrew Zisserman.
\newblock Kinetics dataset.
\newblock In \emph{CVPR}, 2017.

\bibitem[Chen et~al.(2020)Chen, Kornblith, Norouzi, and Hinton]{chen2020simple}
Ting Chen, Simon Kornblith, Mohammad Norouzi, and Geoffrey Hinton.
\newblock A simple framework for contrastive learning of visual representations.
\newblock In \emph{ICML}, 2020.

\bibitem[Fan et~al.(2021)Fan, Xiong, Mangalam, Li, Yan, Malik, and Feichtenhofer]{fan2021multiscale}
Haoqi Fan, Bo Xiong, Karttikeya Mangalam, Yanghao Li, Zhicheng Yan, Jitendra Malik, and Christoph Feichtenhofer.
\newblock Multiscale vision transformers.
\newblock In \emph{CVPR}, 2021.

\bibitem[Fathi et~al.(2011)Fathi, Ren, and Rehg]{gtea}
Alireza Fathi, Xiaofeng Ren, and James~M. Rehg.
\newblock Learning to recognize objects in egocentric activities.
\newblock In \emph{CVPR}, 2011.

\bibitem[Grauman et~al.(2022)Grauman, Westbury, Byrne, Chavis, Furnari, Girdhar, Hamburger, Jiang, Liu, Liu, et~al.]{grauman2022ego4d}
Kristen Grauman, Andrew Westbury, Eugene Byrne, Zachary Chavis, Antonino Furnari, Rohit Girdhar, Jackson Hamburger, Hao Jiang, Miao Liu, Xingyu Liu, et~al.
\newblock Ego4d.
\newblock In \emph{CVPR}, 2022.

\bibitem[Jenamani et~al.(2024)Jenamani, Stabile, Liu, Anwar, Dimitropoulou, and Bhattacharjee]{jenamani2024feel}
Rajat~Kumar Jenamani, Daniel Stabile, Ziang Liu, Abrar Anwar, Katherine Dimitropoulou, and Tapomayukh Bhattacharjee.
\newblock Feel the bite.
\newblock In \emph{HRI}, 2024.

\bibitem[Khosla et~al.(2020)Khosla, Teterwak, Wang, Sarna, Tian, Isola, Maschinot, Liu, and Krishnan]{10.5555/3495724.3497291}
Prannay Khosla, Piotr Teterwak, Chen Wang, Aaron Sarna, Yonglong Tian, Phillip Isola, Aaron Maschinot, Ce Liu, and Dilip Krishnan.
\newblock Supervised contrastive learning.
\newblock In \emph{NeurIPS}, 2020.

\bibitem[Koupaee and Wang(2018)]{koupaee2018wikihow}
Mahnaz Koupaee and William~Yang Wang.
\newblock Wikihow: A large scale text summarization dataset.
\newblock \emph{arXiv}, 2018.

\bibitem[Kung et~al.(2024)Kung, Lu, Tsai, and Chen]{Kung_2024_CVPR}
Chi-Hsi Kung, Shu-Wei Lu, Yi-Hsuan Tsai, and Yi-Ting Chen.
\newblock Action-slot.
\newblock In \emph{CVPR}, 2024.

\bibitem[Kung et~al.(2025)Kung, Ramirez, Ha, Chen, Crandall, and Tsai]{kung2025changed}
Chi-Hsi Kung, Frangil Ramirez, Juhyung Ha, Yi-Ting Chen, David Crandall, and Yi-Hsuan Tsai.
\newblock What changed and what could have changed? state-change counterfactuals for procedure-aware video representation learning.
\newblock \emph{arXiv preprint}, 2025.

\bibitem[Lee et~al.(2024)Lee, Lu, Zhang, Hoai, and Elhamifar]{egoper}
Shih-Po Lee, Zijia Lu, Zekun Zhang, Minh Hoai, and Ehsan Elhamifar.
\newblock Error detection in egocentric procedural task videos.
\newblock In \emph{CVPR}, 2024.

\bibitem[Liang et~al.(2022)Liang, Wang, Zhou, and Yang]{liang2022var}
Chen Liang, Wenguan Wang, Tianfei Zhou, and Yi Yang.
\newblock Visual abductive reasoning.
\newblock In \emph{CVPR}, 2022.

\bibitem[Lin et~al.(2022{\natexlab{a}})Lin, Wang, Soldan, Wray, Yan, Xu, Gao, Tu, Zhao, Kong, et~al.]{kevin2022egovlp}
Kevin~Qinghong Lin, Alex~Jinpeng Wang, Mattia Soldan, Michael Wray, Rui Yan, Eric~Zhongcong Xu, Difei Gao, Rongcheng Tu, Wenzhe Zhao, Weijie Kong, et~al.
\newblock Egocentric video-language pretraining.
\newblock \emph{arXiv preprint}, 2022{\natexlab{a}}.

\bibitem[Lin et~al.(2022{\natexlab{b}})Lin, Petroni, Bertasius, Rohrbach, Chang, and Torresani]{lin2022learning}
Xudong Lin, Fabio Petroni, Gedas Bertasius, Marcus Rohrbach, Shih-Fu Chang, and Lorenzo Torresani.
\newblock Learning to recognize procedural activities with distant supervision.
\newblock In \emph{CVPR}, 2022{\natexlab{b}}.

\bibitem[Mavroudi et~al.(2023)Mavroudi, Afouras, and Torresani]{mavroudi2023learning}
Effrosyni Mavroudi, Triantafyllos Afouras, and Lorenzo Torresani.
\newblock Learning to ground instructional articles in videos through narrations.
\newblock In \emph{ICCV}, 2023.

\bibitem[Miech et~al.(2019)Miech, Zhukov, Alayrac, Tapaswi, Laptev, and Sivic]{miech2019howto100m}
Antoine Miech, Dimitri Zhukov, Jean-Baptiste Alayrac, Makarand Tapaswi, Ivan Laptev, and Josef Sivic.
\newblock Howto100m.
\newblock In \emph{ICCV}, 2019.

\bibitem[Miech et~al.(2020)Miech, Alayrac, Smaira, Laptev, Sivic, and Zisserman]{miech2020end}
Antoine Miech, Jean-Baptiste Alayrac, Lucas Smaira, Ivan Laptev, Josef Sivic, and Andrew Zisserman.
\newblock End-to-end learning of visual representations from uncurated instructional videos.
\newblock In \emph{CVPR}, 2020.

\bibitem[Niu et~al.(2024)Niu, Guo, Chen, Lin, and Chang]{niu2024schema}
Yulei Niu, Wenliang Guo, Long Chen, Xudong Lin, and Shih-Fu Chang.
\newblock Schema: State changes matter for procedure planning in instructional videos.
\newblock \emph{arXiv}, 2024.

\bibitem[Radford et~al.(2021)Radford, Kim, Hallacy, Ramesh, Goh, Agarwal, Sastry, Askell, Mishkin, Clark, et~al.]{radford2021learning}
Alec Radford, Jong~Wook Kim, Chris Hallacy, Aditya Ramesh, Gabriel Goh, Sandhini Agarwal, Girish Sastry, Amanda Askell, Pamela Mishkin, Jack Clark, et~al.
\newblock Learning transferable visual models from natural language supervision.
\newblock In \emph{ICLR}, 2021.

\bibitem[Ren et~al.(2025)Ren, Sundaresan, Sadigh, Choudhury, and Bohg]{ren2025motion}
Juntao Ren, Priya Sundaresan, Dorsa Sadigh, Sanjiban Choudhury, and Jeannette Bohg.
\newblock Motion tracks: A unified representation for human-robot transfer in few-shot imitation learning.
\newblock \emph{arXiv preprint}, 2025.

\bibitem[Sou{\v{c}}ek et~al.(2022)Sou{\v{c}}ek, Alayrac, Miech, Laptev, and Sivic]{souvcek2022look}
Tom{\'a}{\v{s}} Sou{\v{c}}ek, Jean-Baptiste Alayrac, Antoine Miech, Ivan Laptev, and Josef Sivic.
\newblock Look for the change: Learning object states and state-modifying actions from untrimmed web videos.
\newblock In \emph{CVPR}, 2022.

\bibitem[Sou{\v{c}}ek et~al.(2024)Sou{\v{c}}ek, Damen, Wray, Laptev, and Sivic]{souvcek2023genhowto}
Tom{\'a}{\v{s}} Sou{\v{c}}ek, Dima Damen, Michael Wray, Ivan Laptev, and Josef Sivic.
\newblock Genhowto.
\newblock \emph{CVPR}, 2024.

\bibitem[Srinivasan et~al.(2021)Srinivasan, Raman, Chen, Bendersky, and Najork]{srinivasan2021wit}
Krishna Srinivasan, Karthik Raman, Jiecao Chen, Michael Bendersky, and Marc Najork.
\newblock Wit.
\newblock In \emph{Proceedings of the 44th international ACM SIGIR conference on research and development in information retrieval}, pages 2443--2449, 2021.

\bibitem[Tong et~al.(2022)Tong, Song, Wang, and Wang]{tong2022videomae}
Zhan Tong, Yibing Song, Jue Wang, and Limin Wang.
\newblock Videomae: Masked autoencoders are data-efficient learners for self-supervised video pre-training.
\newblock \emph{NeurIPS}, 2022.

\bibitem[Wray et~al.(2021)Wray, Doughty, and Damen]{Wray_2021_CVPR}
Michael Wray, Hazel Doughty, and Dima Damen.
\newblock On semantic similarity in video retrieval.
\newblock In \emph{CVPR}, 2021.

\bibitem[Xiao et~al.(2022)Xiao, Kundu, Tighe, and Modolo]{xiao2022hierarchical}
Fanyi Xiao, Kaustav Kundu, Joseph Tighe, and Davide Modolo.
\newblock Hierarchical self-supervised representation learning for movie understanding.
\newblock In \emph{CVPR}, 2022.

\bibitem[Xue and Grauman(2023)]{xue2023learning}
Zihui Xue and Kristen Grauman.
\newblock Learning fine-grained view-invariant representations from unpaired ego-exo videos via temporal alignment.
\newblock In \emph{NeurIPS}, 2023.

\bibitem[Yi et~al.(2021)Yi, Wen, and Jiang]{chinayi_ASformer}
Fangqiu Yi, Hongyu Wen, and Tingting Jiang.
\newblock Asformer: Transformer for action segmentation.
\newblock In \emph{BMVC}, 2021.

\bibitem[Zhang et~al.(2021)Zhang, Min, Yang, Liu, Jiang, and Rui]{zhang2021if}
Tianyu Zhang, Weiqing Min, Jiahao Yang, Tao Liu, Shuqiang Jiang, and Yong Rui.
\newblock What if we could not see? counterfactual analysis for egocentric action anticipation.
\newblock In \emph{IJCAI}, 2021.

\bibitem[Zhong et~al.(2023)Zhong, Yu, Bai, Li, Yan, and Li]{zhong2023learning}
Yiwu Zhong, Licheng Yu, Yang Bai, Shangwen Li, Xueting Yan, and Yin Li.
\newblock Learning procedure-aware video representation from instructional videos and their narrations.
\newblock In \emph{CVPR}, 2023.

\bibitem[Zhou et~al.(2023)Zhou, Mart{\'\i}n-Mart{\'\i}n, Kapadia, Savarese, and Niebles]{zhou2023procedure}
Honglu Zhou, Roberto Mart{\'\i}n-Mart{\'\i}n, Mubbasir Kapadia, Silvio Savarese, and Juan~Carlos Niebles.
\newblock Procedure-aware pretraining for instructional video understanding.
\newblock In \emph{CVPR}, 2023.

\end{thebibliography}
}


\end{document}